\documentclass[10pt, twocolumn, letterpaper]{article}

\usepackage[pagenumbers]{cvpr} % To force page numbers, e.g. for an arXiv version

\usepackage{graphicx}
\usepackage{capt-of}
\usepackage{amsmath,amssymb,amsfonts,dsfont,bm,bbm,mathrsfs,pifont}
\usepackage[ruled,linesnumbered]{algorithm2e}
\usepackage{listings}
\usepackage{booktabs,multirow,adjustbox}
\usepackage{float}
\usepackage[pagebackref,breaklinks,colorlinks]{hyperref}
\usepackage[capitalize]{cleveref}  % Should be loaded after 'hyperref', and works perfectly with 'subfigure'.
\crefname{section}{Sec.}{Secs.}
\Crefname{section}{Section}{Sections}
\crefname{table}{Tab.}{Tabs.}
\Crefname{table}{Table}{Tables}
\crefname{figure}{Fig.}{Figs.}
\Crefname{figure}{Figure}{Figures}
\crefname{equation}{Eq.}{Eqs.}
\Crefname{equation}{Equation}{Equations}
\crefname{algorithm}{Algorithm}{Algorithms}
\def\cvprPaperID{5131}
\def\confName{CVPR}
\def\confYear{2022}

\begin{document}

%%%% Title
\title{Semi-Supervised Semantic Segmentation Using Unreliable Pseudo-Labels}

\author{
    Yuchao Wang$^{1\dag}$ \quad
    Haochen Wang$^{1\dag}$ \quad
    Yujun Shen$^2$ \quad
    Jingjing Fei$^3$ \quad \\
    Wei Li$^3$ \quad
    Guoqiang Jin$^3$ \quad
    Liwei Wu$^3$ \quad
    Rui Zhao$^{1,3\ddagger}$ \quad
    Xinyi Le$^{1*}$ \\[8pt]
    $^1$Shanghai Jiao Tong University \quad
    $^2$The Chinese University of Hong Kong \quad
    $^3$SenseTime Research \quad \\[5pt]
    \small{\texttt{\{44442222, wanghaochen0409, lexinyi\}@sjtu.edu.cn}} \qquad
    \small{\texttt{shenyujun0302@gmail.com}} \\
    \small{\texttt{\{feijingjing1, liwei1, jinguoqiang, wuliwei, zhaorui\}@sensetime.com}}
}

\maketitle

% \twocolumn[
% \begin{@twocolumnfalse}
%     \maketitle
%     %%%% Authors
%     \vspace{-46pt}
%     \begin{center}
%         \large
%         Yuchao Wang$^{1,3\dag}$\quad
%         Haochen Wang$^{1,3\dag}$\quad
%         Yujun Shen$^2$\quad
%         Jingjing Fei$^3$\quad
        
%         Wei Li$^3$\quad
%         Guoqiang Jin$^3$\quad
%         Liwei Wu$^3$\quad
%         Rui Zhao$^3$\quad
%         Xinyi Le$^{1\ddag}$
        
%         \vspace{3pt}
%         $^1$Shanghai Jiao Tong University\quad
%         $^2$The Chinese University of Hong Kong\quad
%         $^3$SenseTime Research\quad
        
%         \vspace{3pt}
%         \small
%         \texttt{\{44442222, wanghaochen0409, lexinyi\}@sjtu.edu.cn\qquad sy116@ie.cuhk.edu.hk}
            
%         \texttt{\{feijingjing1, liwei1, jinguoqiang, wuliwei, zhaorui\}@sensetime.com}
%         \vspace{3pt}
%     \end{center}
%     \vspace{24pt}
% \end{@twocolumnfalse}
% ]

%%%% Abstract
\begin{abstract}
The crux of semi-supervised semantic segmentation is to assign adequate pseudo-labels to the pixels of unlabeled images.
A common practice is to select the highly confident predictions as the pseudo ground-truth, but it leads to a problem that most pixels may be left unused due to their unreliability.
We argue that every pixel matters to the model training, even its prediction is ambiguous.
Intuitively, an unreliable prediction may get confused among the top classes (\textit{i.e.}, those with the highest probabilities), however, it should be confident about the pixel not belonging to the remaining classes.
Hence, such a pixel can be convincingly treated as a negative sample to those most unlikely categories.
Based on this insight, we develop an effective pipeline to make sufficient use of unlabeled data.
Concretely, we separate reliable and unreliable pixels via the entropy of predictions, push each unreliable pixel to a category-wise queue that consists of negative samples, and manage to train the model with all candidate pixels.
Considering the training evolution, where the prediction becomes more and more accurate, we adaptively adjust the threshold for the reliable-unreliable partition.
Experimental results on various benchmarks and training settings demonstrate the superiority of our approach over the state-of-the-art alternatives.%
\footnote{
Project: \url{https://haochen-wang409.github.io/U2PL}.

~$^*$Corresponding author. This work is sponsored by National Natural Science Foundation of China (62176152).

~$^\dag$Equal contribution, and this work is done during the internship at SenseTime Research.

~$^\ddagger$Rui Zhao is also with Qing Yuan Research Institute, Shanghai Jiao Tong University.
}

% Semi-supervised semantic segmentation methods utilize a large amount of unlabeled set to make decision boundaries lie on low density regions. \yj{Why do we need this sentence?}
% %
% Most previous methods mainly focus on a) generating or selecting high quality pseudo-labels for unlabeled images; b) re-weighting or re-sampling for long-tailed classes.
% %
% All of them ignore those unreliable but valuable pseudo-labels.
% %
% Our main insight is that those unreliable pixels are just confused in the top 2 or 3 classes, which means they are confident enough for not belonging to the remaining categories.
% %
% Therefore, we present U$^2$PL, a novel pixel-wise contrastive learning based method by \textbf{u}sing \textbf{u}nreliable \textbf{p}seudo \textbf{l}abels.
% %
% Specifically, based on the standard InfoNCE loss, we design a novel strategy for sampling negative
% pairs, making full use of uncertain pixels.
% %
% Besides, we filter high quality pseudo-labels for unlabeled images based on pixel-level entropy map and apply strong data augmentation on unlabeled data.
% %
% Our method outperforms other state-of-the-art semi-supervised semantic segmentation on Cityscapes and Pascal VOC benchmarks under various protocols.
% %
% % Codes will be available at \url{xxxx}.
% %
\end{abstract}

%%%% Section: Introduction
\section{Introduction}\label{sec:intro}

\begin{figure}[t]
    \centering
    \includegraphics[width=1.0\linewidth]{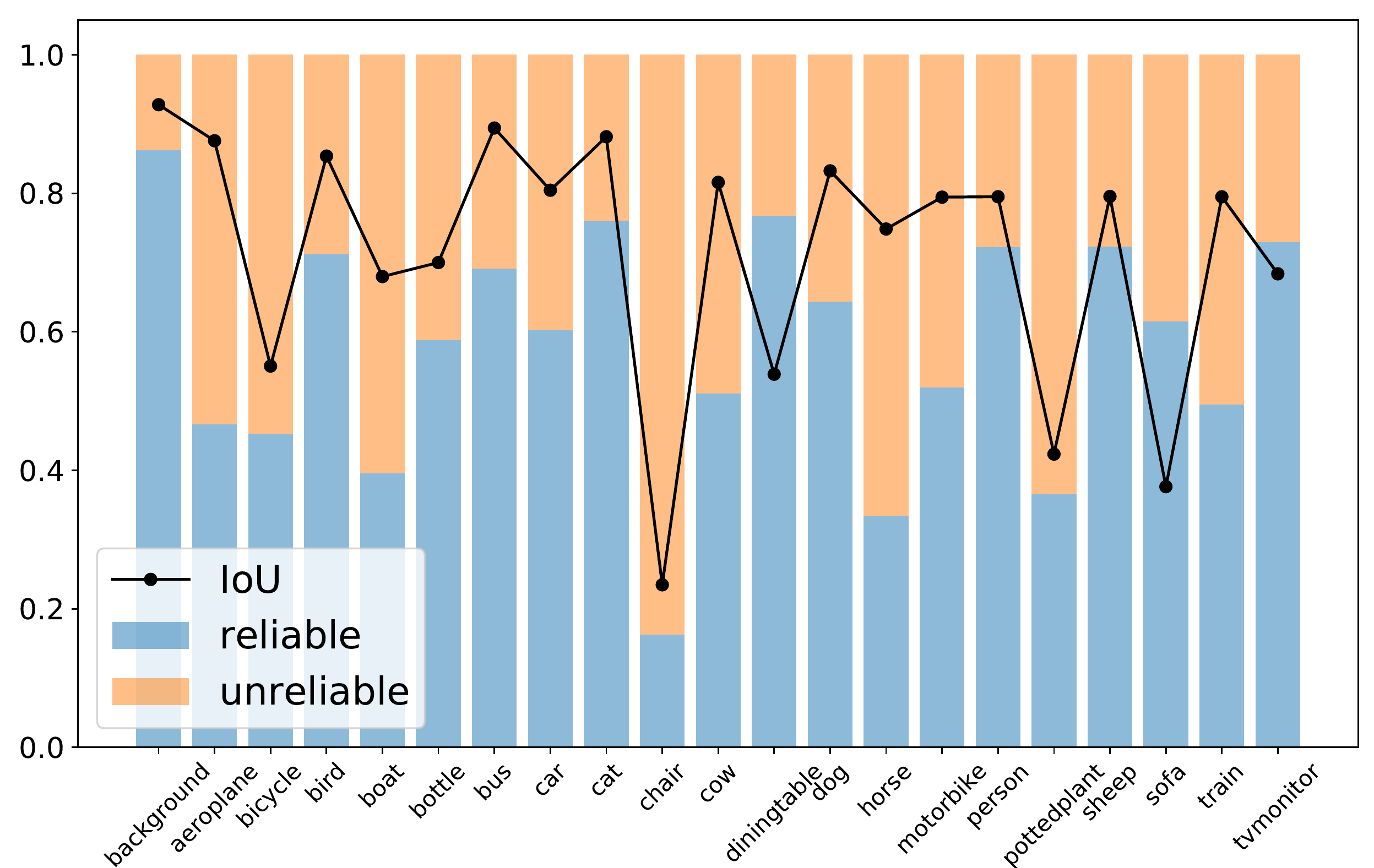}
    \vspace{-18pt}
    \caption{
        \textbf{Category-wise performance and statistics on number of pixels with reliable and unreliable predictions.}
        Model is trained using $732$ labeled images on PASCAL VOC 2012~\cite{voc} and evaluated on the remaining $9,850$ images.
    }
    \label{fig:stats}
    \vspace{-12pt}
\end{figure}

Semantic segmentation is a fundamental task in the computer vision field, and has been significantly advanced along with the rise of deep neural networks~\cite{fcn, unet, pspnet, deeplab}.
Existing supervised approaches rely on large-scale annotated data, which can be too costly to acquire in practice.
%
% remote sensing interpretation and medical image analysis.
%
To alleviate this problem, many attempts~\cite{st++, cps, chen2021semisupervised, alonso2021semi, french2019semi, cct, ael, pc2seg} have been made towards semi-supervised semantic segmentation, which learns a model with only a few labeled samples and numerous unlabeled ones.
Under such a setting, how to adequately leverage the unlabeled data becomes critical.

A typical solution is to assign pseudo-labels to the pixels without annotations.
Concretely, given an unlabeled image, prior arts~\cite{lee2013pseudo, xie2020self} borrow predictions from the model trained on labeled data, and use the pixel-wise prediction as the ``ground-truth'' to in turn boost the supervised model.
To mitigate the problem of confirmation bias~\cite{arazo2020pseudo}, where the model may suffer from incorrect pseudo-labels, existing approaches propose to filter the predictions with their confidence scores~\cite{st++, pseudoseg, zuo2021self, dash}.
In other words, only the highly confident predictions are used as the pseudo-labels, while the ambiguous ones are discarded.

However, one potential problem caused by only using reliable predictions is that some pixels may never be learned in the entire training process.
For example, if the model cannot satisfyingly predict some certain class (\textit{e.g.}, \texttt{chair} in \cref{fig:stats}), it becomes difficult to assign accurate pseudo-labels to the pixels regarding such a class, which may lead to insufficient and categorically imbalanced training.
%
% Such a case especially occurs on some long-tailed categories.
%
From this perspective, we argue that, to make full use of the unlabeled data, every pixel should be properly utilized.

\begin{figure}[t]
    \centering
    \includegraphics[width=1.0\linewidth]{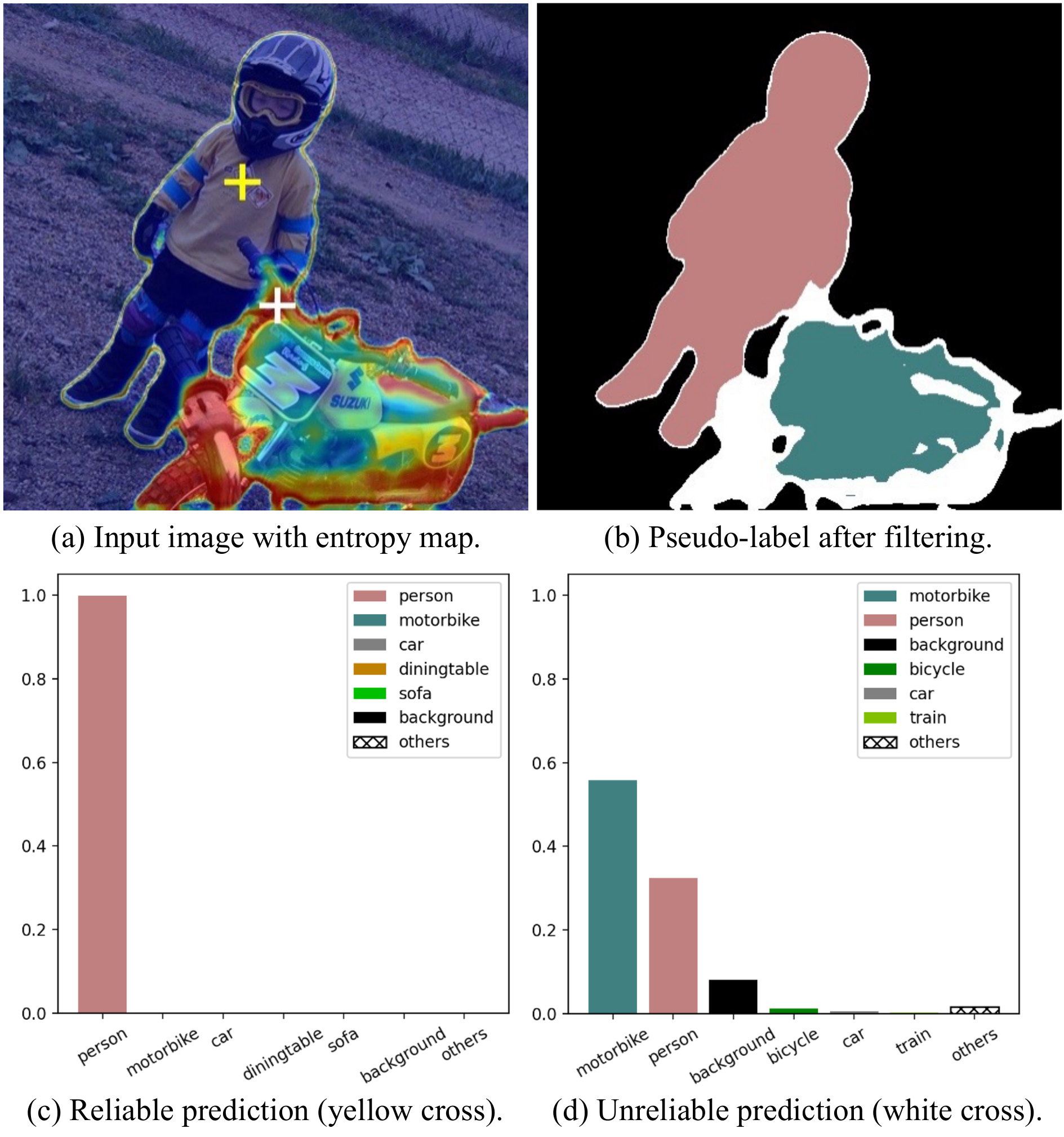}
    \vspace{-18pt}
    \caption{
        \textbf{Illustration on unreliable pseudo-labels.}
        (a) Pixel-wise entropy predicted from an unlabeled image, where low-entropy pixels and high-entropy pixels indicate reliable and unreliable predictions, respectively.
        (b) Pixel-wise pseudo-labels from reliable predictions \textit{only}, where pixels within the white region are not assigned a pseudo-label.
        (c) Category-wise probability of a reliable prediction (\textit{i.e.}, the yellow cross), which is confident enough for supervising the class \texttt{person}.
        (d) Category-wise probability of an unreliable prediction (\textit{i.e.}, the white cross), which hovers between \texttt{motorbike} and \texttt{person}, yet is confident enough of \textit{not} belonging to \texttt{car} and \texttt{train}.
    }
    \label{fig:example}
    \vspace{-12pt}
\end{figure}

As discussed above, directly using the unreliable predictions as the pseudo-labels will cause the performance degradation~\cite{arazo2020pseudo}.
In this paper, we propose an alternative way of Using Unreliable Pseudo-Labels.
We call our framework as U$^2$PL.
First, we observe that, an unreliable prediction usually gets confused among \textit{only a few} classes instead of all classes.
Taking \cref{fig:example} as an instance, the pixel with white cross receives similar probabilities on class \texttt{motorbike} and \texttt{person}, but the model is pretty sure about this pixel \textit{not} belonging to class \texttt{car} and \texttt{train}.
Based on this observation, we reconsider the confusing pixels as the negative samples to those unlikely categories.
Specifically, after getting the prediction from an unlabeled image, we employ the per-pixel entropy as the metric (see \cref{fig:example}\textcolor{red}{a}) to separate all pixels into two groups, \textit{i.e.}, a reliable one and an unreliable one.
All reliable predictions are used to derive positive pseudo-labels, while the pixels with unreliable predictions are pushed into a memory bank, which is full of negative samples.
To avoid all negative pseudo-labels only coming from a subset of categories, we employ a queue for each category.
Such a design ensures that the number of negative samples for each class is balanced.
Meanwhile, considering that the quality of pseudo-labels becomes higher along with the model gets more and more accurate, we come up with a strategy to adaptively adjust the threshold for the partition of reliable and unreliable pixels.

We evaluate the proposed U$^2$PL on PASCAL VOC 2012~\cite{voc} and Cityscapes~\cite{cityscapes} under a wide range of training settings, where our approach surpasses the state-of-the-art competitors.
Furthermore, through visualizing the segmentation results, we find that our method achieves much better performance on those ambiguous regions (\textit{e.g.}, the border between different objects), thanks to our adequate use of the unreliable pseudo-labels.

%%%% Section: Related Work
\section{Related Work}\label{sec:related}

\noindent\textbf{Semi-Supervised Learning} has two typical paradigms: consistency regularization~\cite{bachman2014learning, cct, french2019semi, sajjadi2016regularization, dash} and entropy minimization~\cite{em, chen2021semisupervised}. 
Recently, a more intuitive but effective framework: self-training~\cite{lee2013pseudo}, has become the mainstream.
Several methods~\cite{french2019semi, yuan2021simple, st++} utilize strong data augmentation such as CutOut~\cite{cutout}, CutMix~\cite{cutmix}, and ClassMix~\cite{classmix} based on self-training. 
However, these methods do not pay much attention to the characteristics of semantic segmentation, while our method focuses on those \textit{unreliable pixels} which will be filtered out by most of self-training based methods~\cite{yuan2021simple, st++, ups}.

\noindent\textbf{Pseudo-Labeling} is applied to prevent overfitting to incorrect pseudo-labels when generating predictions of input images from the teacher network~\cite{lee2013pseudo, arazo2020pseudo}.
FixMatch~\cite{fixmatch} utilizes a confidence threshold to select reliable pseudo-labels.
UPS~\cite{ups}, a method based on FixMatch~\cite{fixmatch}, takes model uncertainty and data uncertainty into consideration. 
However, in semi-supervised semantic segmentation, our experiments show including unreliable pixels into training can boost performance.

\noindent\textbf{Model Uncertainty} in computer vision is mostly measured by \textit{Bayesian deep learning approaches}~\cite{der2009aleatory, kendall2017uncertainties, mukhoti2018evaluating}.
In our settings, we do not focus on how to measure uncertainty.
We simply use the entropy of pixel-wise probability distribution to be the metric.

\noindent\textbf{Contrastive Learning} is applied by many successful works in self-supervised learning~\cite{simclrv2, mocov3, byol}.
In semantic segmentation, contrastive learning has become a promising new paradigm~\cite{reco, wang2021exploring, zhao2021contrastive, alonso2021semi, zhou2021c3}. 
%
% Following methods try to go deeper by adopting contrastive learning framework for semi-supervised semantic segmentation task.
% %
% PC${}^2$Seg~\cite{pc2seg} minimizes the mean square error between two positive samples, and introduces several strategy to sample negative pixels. 
% %
% Alonso~\textit{et al.}~\cite{alonso2021semi} utilizes a class-wise memory bank to store representative negative pixels for each class. 
%
However, these methods ignore the common \textit{false negative samples} in semi-supervised segmentation, and unreliable pixels may be wrongly pushed away in contrastive loss.
Discriminating the unlikely categories of unreliable pixels can addresses this problem.

% All methods above select reliable pixels based on confidence or entropy to avoid pushing or pulling wrong pixel pairs when samples are unlabeled. 
% %
% However, in our opinion, pixels with low confidence or high entropy are more worth optimized, which has never been considered due to the risk of being classified into a wrong class in semi-supervised learning.

\begin{figure*}[t]
    \centering
    \includegraphics[width=1\textwidth]{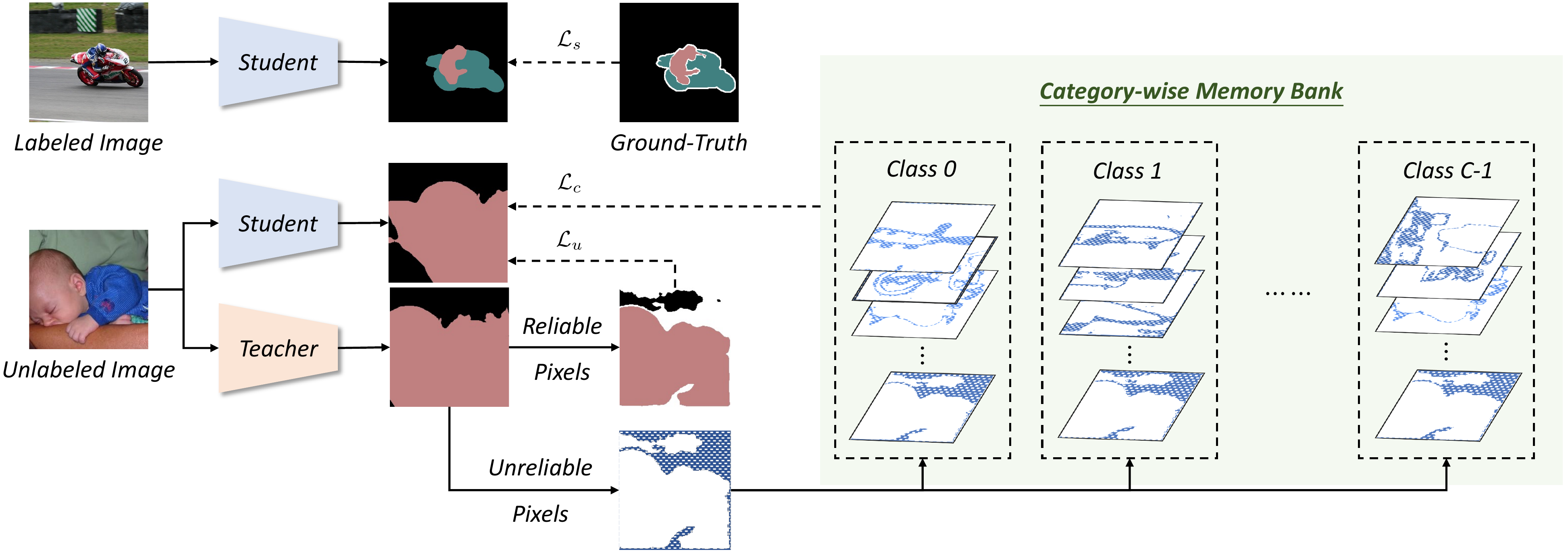}
    \caption{
        \textbf{An overview of our proposed U$^2$PL method.} 
        U$^2$PL contains a student network and a teacher network, where the teacher is momentum-updated with the student.
        Labeled data is directly fed into the student network for supervised training.
        Given an unlabeled image, we first use the teacher model to make a prediction, and then separate the pixels into reliable ones and unreliable ones based on their entropy.
        Such a process is formulated as \cref{eq:pseudo}.
        The reliable predictions are directly used as the pseudo-labels to advise the student, while each unreliable prediction is pushed into a category-wise memory bank.
        Pixels in each memory bank are regarded as the negative samples to the corresponding class, which is formulated as \cref{eq:contraloss}.
    }
    \label{fig:pipeline}
    \vspace{-15pt}
\end{figure*}

\noindent\textbf{Negative Learning} aims at decreasing the risk of incorrect information by  lowering the probability of negative samples~\cite{kim2019nlnl, tokunaga2020negative, ups, kim2021joint}, but those negative samples are selected with high confidence.
In other words, these methods still utilizes pixels with reliable predictions.
By contrast, we propose to make sufficient use of those unreliable predictions for learning instead of filtering them out.

%%%% Section: Method

\section{Method}\label{sec:method}

In this section, we establish our problem mathematically and give an overview of our proposed method in \cref{sec:overview} first. Our strategies about filtering reliable pseudo-labels are introduced in \cref{sec:pseudo}. Finally, we describe how to use unreliable pseudo-labels in \cref{sec:unreliable}.

\subsection{Overview}\label{sec:overview}

Given a labeled set $\mathcal{D}_l=\left\{(\mathbf{x}_i^l, \mathbf{y}_i^l)\right\}_{i=1}^{N_l}$ and a much larger unlabeled set $\mathcal{D}_u=\left\{\mathbf{x}_i^u\right\}_{i=1}^{N_u}$,
our goal is to train a semantic segmentation model by leveraging both a large amount of unlabeled data and a smaller set of labeled data.

\cref{fig:pipeline} gives an overview of U$^2$PL, which follows the typical self-training framework with two models of the same architecture, named teacher and student respectively.
The two models differ only when updating their weights.
The student model's weights $\theta_s$ are updated consistent with the common practice while the teacher model's weights $\theta_t$ are exponential moving average (EMA) updated by the student model's weights.
Each model consists of a CNN-based encoder $h$, a decoder with a segmentation head $f$, and a representation head $g$.
At each training step, we equally sample $B$ labeled images $\mathcal{B}_l$ and $B$ unlabeled images $\mathcal{B}_u$.
For every labeled image, our goal is to minimize the standard cross-entropy loss in \cref{eq:suploss}.
As for each unlabeled image, we first take it into the teacher model and get predictions.
Then, based on pixel-level entropy, we ignore unreliable pixel-level pseudo-labels when computing unsupervised loss in \cref{eq:unsloss}.
This part will be introduced in section \cref{sec:pseudo} in detail.
Finally, we use the contrastive loss to make full use of the unreliable pixels excluded in the unsupervised loss, which will be introduced in \cref{sec:unreliable}.

Our optimization target is to minimize the overall loss, which can be formulated as:
\begin{equation}
    \mathcal{L} = \mathcal{L}_s + \lambda_u \mathcal{L}_u + \lambda_c \mathcal{L}_c,
\end{equation}
where $\mathcal{L}_s$ and $\mathcal{L}_u$ represent supervised loss and unsupervised loss applied on labeled images and unlabeled images respectively, and $\mathcal{L}_c$ is the contrastive loss to make full use of unreliable pseudo-labels. 
$\lambda_u$ and $\lambda_c$ are weights of unsupervised loss and contrastive loss respectively.
Both $\mathcal{L}_s$ and $\mathcal{L}_u$ are cross-entropy (CE) loss:
\begin{equation}
\label{eq:suploss}
    \mathcal{L}_s = \frac{1}{|\mathcal{B}_l|} \sum_{(\mathbf{x}_i^l, \mathbf{y}_i^l) \in \mathcal{B}_l} \ell_{ce}(f\circ h(\mathbf{x}_i^l; \theta), \mathbf{y}_i^l),
\end{equation}
\begin{equation}
\label{eq:unsloss}
    \mathcal{L}_u = \frac{1}{|\mathcal{B}_u|} \sum_{\mathbf{x}_i^u \in \mathcal{B}_u} \ell_{ce}(f\circ h(\mathbf{x}_i^u; \theta), \hat{\mathbf{y}}_i^u),
\end{equation}
where $\mathbf{y}_i^l$ represents the hand-annotated mask label for the $i$-th labeled image, and $\hat{\mathbf{y}}_i^u$ is the pseudo-label for the $i$-th unlabeled image.
$f\circ h$ is the composition function of $h$ and $f$, which means the images are first fed into $h$ and then $f$ to get segmentation results.
$\mathcal{L}_c$ is the pixel-level InfoNCE~\cite{infonce} loss defined as:
\begin{equation}
\label{eq:contraloss}
\begin{aligned}
    \mathcal{L}_c = &- \frac{1}{C\times M}\sum_{c=0}^{C-1} \sum_{i=1}^M \\
    &\log\left[ \frac{e^{\langle \mathbf{z}_{ci}, \mathbf{z}_{ci}^{+}\rangle / \tau}}{e^{\langle \mathbf{z}_{ci}, \mathbf{z}_{ci}^{+}\rangle / \tau} + \sum_{j=1}^N e^{\langle \mathbf{z}_{ci}, \mathbf{z}_{cij}^{-}\rangle / \tau}} \right],
\end{aligned}
\end{equation}
where $M$ is the total number of anchor pixels, and $\mathbf{z}_{ci}$ denotes the representation of the $i$-th anchor of class $c$.
Each anchor pixel is followed with a positive sample and $N$ negative samples, whose representations are $\mathbf{z}_{ci}^{+}$ and $\mathbf{z}_{cij}^{-}$ respectively.
Note that $\mathbf{z}=g\circ h (\mathbf{x})$ is the output of the representation head.
$\langle \cdot, \cdot \rangle$ is the cosine similarity between features from two different pixels, whose range is limited between $-1$ to $1$, hence the need of temperature $\tau$.
Following~\cite{reco}, we set $M=50$, $N=256$ and $\tau=0.5$.

% Following~\cite{meanteacher}, we use Mean Teacher as our basic framework, which consists of a teacher network and a student network.
% %
% Parameters of the teacher are updated by an exponential moving average (EMA) manner:
% %
% \begin{equation}
% \label{eq:ema}
%     \theta_T = \omega \theta_T + (1 - \omega) \theta_S,
% \end{equation}
% %
% where $\theta_T$ and $\theta_S$ represent the parameters of the teacher model and the student model respectively.
% %
% $\omega\in[0, 1)$ is a momentum coefficient, and only parameters $\theta_S$ are updated by back-propagation.

\subsection{Pseudo-Labeling}
\label{sec:pseudo}
To avoid overfitting incorrect pseudo-labels, we utilize entropy of every pixel's probability distribution to filter high quality pseudo-labels for further supervision.
Specifically, we denote $\mathbf{p}_{ij}\in\mathbb{R}^C$ as the softmax probabilities generated by the segmentation head of the teacher model for the $i$-th unlabeled image at pixel $j$, where $C$ is the number of classes.
Its entropy is computed by:
\begin{equation}
    \mathcal{H}(\mathbf{p}_{ij}) = -\sum_{c=0}^{C-1} p_{ij}(c)\log p_{ij}(c),
\end{equation}
where $p_{ij}(c)$ is the value of $\mathbf{p}_{ij}$ at $c$-th dimension.

Then, we define pixels whose entropy on top $\alpha_t$ as unreliable pseudo-labels at training epoch $t$. 
Such unreliable pseudo-labels are not qualified for supervision.
Therefore, we define the pseudo-label for the $i$-th unlabeled image at pixel $j$ as:
\begin{equation}
\label{eq:pseudo}
    \hat{y}_{ij}^u = \left\{
    \begin{aligned}
        &\arg\max_c p_{ij}(c), &&\mathrm{if}\ \mathcal{H}(\mathbf{p}_{ij}) < \gamma_t, \\
        &\mathrm{ignore}, &&\mathrm{otherwise},
    \end{aligned}
    \right.
\end{equation}
where $\gamma_t$ represents the entropy threshold at $t$-th training step. 
We set $\gamma_t$ as the quantile corresponding to $\alpha_t$, \textit{i.e.}, $\gamma_t$=\texttt{np.percentile(H.flatten(),100*(1-$\alpha_t$))}, where \texttt{H} is per-pixel entropy map.
We adopt the following adjustment strategies in the pseudo-labeling process for better performance.

\noindent \textbf{Dynamic Partition Adjustment.} During the training procedure, the pseudo-labels tend to be reliable gradually.
Base on this intuition, we adjust unreliable pixels' proportion $\alpha_t$ with linear strategy every epoch:
\begin{equation}
\label{eq:dpa}
    \alpha_t = \alpha_0 \cdot \left(1 - \frac{t}{\mathrm{total\ epoch}}\right),
\end{equation}
where $\alpha_0$ is the initial proportion and is set to $20\%$, and $t$ is the current training epoch.

\noindent \textbf{Adaptive Weight Adjustment.} After obtaining reliable pseudo-labels, we involve them in the unsupervised loss in \cref{eq:unsloss}. 
The weight $\lambda_u$ for this loss is defined as the reciprocal of the percentage of pixels with entropy smaller than threshold $\gamma_t$ in the current mini-batch multiplied by a base weight $\eta$:
\begin{equation}
\label{eq:awa}
    \lambda_u = \eta \cdot \frac{|\mathcal{B}_u|\times H\times W}{\sum_{i=1}^{|\mathcal{B}_u|} \sum_{j=1}^{H\times W} \mathbbm{1}\left[\hat{y}_{ij}^u \neq \mathrm{ignore}\right]},
\end{equation}
where $\mathbbm{1}(\cdot)$ is the indicator function and $\eta$ is set to $1$.

\subsection{Using Unreliable Pseudo-Labels}
\label{sec:unreliable}

In semi-supervised learning tasks, discarding unreliable pseudo-labels or reducing their weights is widely used to prevent degradation of model's performance~\cite{pseudoseg, st++, fixmatch, xie2020self}. 
We follow this intuition by filtering out unreliable pseudo-labels based on \cref{eq:pseudo}.

However, such contempt for unreliable pseudo-labels may result in information loss. 
It is obvious that unreliable pseudo-labels can provide information for better discrimination.
For example, the white cross in \cref{fig:example}, is a typical unreliable pixel. 
Its distribution demonstrates model's uncertainty to distinguish between class \texttt{person} and class \texttt{motorbike}.
However, this distribution also demonstrates model's certainty to not to discriminate this pixel as class \texttt{car}, class \texttt{train}, class \texttt{bicycle} and so on.
Such characteristic gives us the main insight to propose our U$^2$PL to use unreliable pseudo-labels for semi-supervised semantic segmentation.

U$^2$PL, with a goal to use the information of unreliable pseudo-labels for better discrimination, coincides with recent popular contrastive learning paradigm in distinguishing representation.
But due to the lack of labeled images in semi-supervised semantic segmentation tasks, our U$^2$PL is built on more complicated strategies.
U$^2$PL has three components, named anchor pixels, positive candidates and negative candidates. 
These components are obtained in a sampling manner from certain sets to alleviate huge computational cost.
Next, we will introduce how to selecting: (a) anchor pixels (queries); (b) positive samples for each anchor; (c) negative samples for each anchor.

\noindent\textbf{Anchor Pixels.}
During training, we sample anchor pixels (queries) for each class that appears in the current mini batch.
We denote the set of features of all labeled candidate anchor pixels for class $c$ as $\mathcal{A}_c^l$, 
\begin{equation}
\label{eq:Al}
    \mathcal{A}_c^l = \left\{
    \mathbf{z}_{ij} \mid y_{ij}=c, p_{ij}(c) > \delta_p
    \right\},
\end{equation}
where $y_{ij}$ is the ground-truth for the $j$-th pixel of labeled image $i$, and $\delta_p$ denotes the positive threshold for a particular class and is set to $0.3$ following~\cite{reco}. 
$\mathbf{z}_{ij}$ means the representation of the $j$-th pixel of labeled image $i$.
For unlabeled data, counterpart $\mathcal{A}_c^u$ can be computed as:
\begin{equation}
\label{eq:Au}
    \mathcal{A}_c^u = \left\{
    \mathbf{z}_{ij} \mid \hat{y}_{ij}=c, p_{ij}(c) > \delta_p
    \right\}.
\end{equation}
It is similar to $\mathcal{A}_c^l$, and the only difference is that we use pseudo-label $\hat{y}_{ij}$ based on \cref{eq:pseudo} rather than hand-annotated label, which implies that qualified anchor pixels are reliable, \textit{i.e.}, $\mathcal{H}(\mathbf{p}_{ij}) \leq \gamma_t$.
Therefore, for class $c$, the set of all qualified anchors is
\begin{equation}
\label{eq:ac}
    \mathcal{A}_c=\mathcal{A}_c^l\cup\mathcal{A}_c^u.
\end{equation}

\noindent\textbf{Positive Samples.}
The positive sample is the same for all anchors from the same class.
It is the center of all possible anchors:
\begin{equation}
\label{eq:positive}
    \mathbf{z}_c^{+} = \frac{1}{|\mathcal{A}_c|} \sum_{\mathbf{z}_c \in \mathcal{A}_c} \mathbf{z}_c.
\end{equation}

\noindent\textbf{Negative Samples.}
We define a binary variable $n_{ij}(c)$ to identify whether the $j$-th pixel of image $i$ is qualified to be negative samples of class $c$.
\begin{equation}
    n_{ij}(c) = \left\{
    \begin{aligned}
        &n_{ij}^l(c), &\mathrm{if\ image\ } i \mathrm{\ is\ labeled}, \\
        &n_{ij}^u(c), &\mathrm{otherwise},
    \end{aligned}
    \right.
\end{equation}
where $n_{ij}^l(c)$ and $n_{ij}^u(c)$ are indicators of whether the $j$-th pixel of labeled and unlabeled image $i$ is qualified to be negative samples of class $c$ respectively.

For $i$-th labeled image, a qualified negative sample for class $c$ should be: (a) not belonging to class $c$; (b) difficult to distinguish between class $c$ and its ground-truth category.
Therefore, we introduce the pixel-level category order $\mathcal{O}_{ij}=\texttt{argsort}(\mathbf{p}_{ij})$. %
Obviously, we have $\mathcal{O}_{ij}(\arg\max\mathbf{p}_{ij})=0$ and $\mathcal{O}_{ij}(\arg\min\mathbf{p}_{ij})=C-1$.
\begin{equation}
    n_{ij}^l(c) = \mathbbm{1}\left[ y_{ij}\neq c \right] \cdot \mathbbm{1}\left[ 0 \leq \mathcal{O}_{ij}(c) < r_l \right],
\end{equation}
where $r_l$ is the low rank threshold and is set to $3$.
The two indicators reflect feature (a) and (b) respectively.

For $i$-th unlabeled image, a qualified negative sample for class $c$ should: (a) be unreliable; (b) probably not belongs to class $c$; (c) not belongs to most unlikely classes.
Similarly, we also use $\mathcal{O}_{ij}$ to define $n_{ij}^u(c)$:
\begin{equation}
    n_{ij}^u(c) = \mathbbm{1}\left[ \mathcal{H}(\mathbf{p}_{ij}) > \gamma_t \right] \cdot \mathbbm{1}\left[ r_l \leq \mathcal{O}_{ij}(c) < r_h \right],
\end{equation}
where $r_h$ is the high rank threshold and is set to $20$.
Finally, the set of negative samples of class $c$ is
\begin{equation}
\label{eq:negative}
    \mathcal{N}_c = \left\{ \mathbf{z}_{ij} \mid n_{ij}(c) = 1 \right\}.
\end{equation}

\noindent\textbf{Category-wise Memory Bank.}
Due to the long tail phenomenon of the dataset, negative candidates in some particular categories are extremely limited in a mini-batch. 
In order to maintain a stable number of negative samples, we use category-wise memory bank $\mathcal{Q}_c$ (FIFO queue) to store the negative samples for class $c$.

Finally, the whole process to use unreliable pseudo-labels is shown in Algorithm \ref{algo:u2pl}.
All features of anchors are attach to gradient, come from student hence,
while features of positive and negative samples are from teacher.

\begin{algorithm}[t]
  \SetAlgoLined

  Initialize $\mathcal{L} \leftarrow 0$\;
  Sample labeled images $\mathcal{B}_l$ and unlabeled images $\mathcal{B}_u$\;
  
  \For{$\mathbf{x}_i \in \mathcal{B}_l\cup\mathcal{B}_u$}{
    Get probabilities: $\mathbf{p}_i \leftarrow f\circ h(\mathbf{x}_i;\theta_t)$\;
    Get representations: $\mathbf{z}_i \leftarrow g\circ h(\mathbf{x}_i;\theta_s)$\;
    
    \For{$c \leftarrow 0$ \KwTo $C-1$}{
      Get anchors $\mathcal{A}_c$ based on \cref{eq:ac}\;
      Sample $M$ anchors: $\mathcal{B}_A \leftarrow $ \texttt{sample} $(\mathcal{A}_c)$\;
      
      Get negatives $\mathcal{N}_c$ based on \cref{eq:negative}\;
      Push $\mathcal{N}_c$ into memory bank $\mathcal{Q}_c$\; 
      Pop oldest ones out of $\mathcal{Q}_c$ if necessary\;
      Sample $N$ negatives: $\mathcal{B}_N \leftarrow $ \texttt{sample} $(\mathcal{Q}_c)$\;
      
      Get $\mathbf{z}^{+}$ based on \cref{eq:positive}\;
      
      $\mathcal{L} \leftarrow \mathcal{L} + \ell(\mathcal{B}_A,\mathcal{B}_N, \mathbf{z}^{+})$ based on \cref{eq:contraloss}\;
    }
  }
  
  \KwOut{contrastive loss $\mathcal{L}_c \leftarrow \frac{1}{|\mathcal{B}|\times C} \mathcal{L}$}
  \caption{Using Unreliable Pseudo-Labels}
  \label{algo:u2pl}
\end{algorithm}

%%%% Section: Experiments
\section{Experiments}\label{sec:exp}

\subsection{Setup}
\noindent\textbf{Datasets.} 
PASCAL VOC 2012~\cite{voc} Dataset is a standard semantic segmentation benchmark with 20 semantic classes of objects and 1 class of background.
The training set and the validation set include $1,464$ and $1,449$ images respectively.
Following~\cite{ael, st++, cps}, we use SBD~\cite{sbd} as the augmented set with $9,118$ additional training images.
Since the SBD~\cite{sbd} dataset is coarsely annotated, PseudoSeg~\cite{pseudoseg} takes only the standard $1,464$ images as the whole labeled set, while other methods~\cite{cps, ael} take all $10,582$ images as candidate labeled data.
Therefore, we evaluate our method on both the \textit{classic} set ($1,464$ candidate labeled images) and the \textit{blender} set ($10,582$ candidate labeled images).
Cityscapes~\cite{cityscapes}, a dataset designed for urban scene understanding, consists of $2,975$ training images with fine-annotated masks and $500$ validation images.
For each dataset, we compare U$^2$PL with other methods under $1/2$, $1/4$, $1/8$, and $1/16$ partition protocols.

\begin{table*}[t]
\centering
\caption{
Comparison with state-of-the-art methods on \textit{classic} \textbf{PASCAL VOC 2012} \texttt{val} set under different partition protocols. 
The labeled images are selected from the original VOC \texttt{train} set, which consists of $1,464$ samples in total.
The fractions denote the percentage of labeled data used for training, followed by the actual number of images.
All the images from SBD~\cite{sbd} are regarded as unlabeled data.
``SupOnly'' stands for supervised training without using any unlabeled data.
\dag\ means we reproduce the approach.
}
\label{tab:classic}
\setlength{\tabcolsep}{13pt}
\vspace{-5pt}
\begin{tabular}{l|lllll}
\toprule
Method & 
1/16 (92) & 1/8 (183) & 1/4 (366) & 1/2 (732) & Full (1464) \\
\midrule
SupOnly & 
45.77 & 54.92 & 65.88 & 71.69 &72.50 \\
\midrule
MT$^\dag$~\cite{meanteacher} & 
51.72 & 58.93 & 63.86 & 69.51 & 70.96 \\
CutMix$^\dag$~\cite{french2019semi} & 
52.16 & 63.47 & 69.46 & 73.73 & 76.54 \\
PseudoSeg~\cite{pseudoseg} & 
57.60 & 65.50 & 69.14 & 72.41 & 73.23 \\
PC${}^2$Seg~\cite{pc2seg} & 
57.00 & 66.28 & 69.78 & 73.05 & 74.15 \\
\midrule
U$^2$PL (w/ CutMix) & 
\textbf{67.98} \footnotesize{(\textcolor{blue}{$+$15.82})} & 
\textbf{69.15} \footnotesize{(\textcolor{blue}{$+$5.68})} & 
\textbf{73.66} \footnotesize{(\textcolor{blue}{$+$4.20})} & 
\textbf{76.16} \footnotesize{(\textcolor{blue}{$+$2.43})} & 
\textbf{79.49} \footnotesize{(\textcolor{blue}{$+$2.95})} \\
\bottomrule
\end{tabular}
\vspace{-5pt}
\end{table*}

\noindent\textbf{Network Structure.} 
We use ResNet-101~\cite{resnet} pre-trained on ImageNet~\cite{imagenet} as the backbone and DeepLabv3+~\cite{deeplabv3p} as the decoder.
Both of the segmentation head and the representation head consists of two \texttt{Conv-BN-ReLU} blocks, where both blocks preserve the feature map resolution and the first block halves the number of channels.
The segmentation head can be seen as a pixel-level classifier, mapping the $512$ dimensional features output from ASPP module into $C$ classes.
The representation head maps the same features into $256$ dimensional representation space.

\noindent\textbf{Evaluation.} 
Following previous methods~\cite{ael, pc2seg, cct, french2019semi}, the images are center cropped into a fixed resolution for PASCAL VOC 2012.
For Cityscapes, previous methods apply slide window evaluation, so do we.
Then we adopt the mean of Intersection over Union (mIoU) as the metric to evaluate these cropped images.
All results are measured on the \texttt{val} set on both Cityscapes~\cite{cityscapes} and PASCAL VOC 2012~\cite{voc}.
Ablation studies are conducted on the \textit{blender} PASCAL VOC 2012~\cite{voc} \texttt{val} set under $1/4$ and $1/8$ partition protocol.

\noindent\textbf{Implementation Details.}
For the training on the \textit{blender} and \textit{classic} PASCAL VOC 2012 dataset, we use stochastic gradient descent (SGD) optimizer with initial learning rate $0.001$,  weight decay as $0.0001$, crop size as $513\times513$, batch size as $16$ and training epochs as $80$.
For the training on the Cityscapes dataset, we also use stochastic gradient descent (SGD) optimizer with initial learning rate $0.01$,  weight decay as $0.0005$, crop size as $769\times769$, batch size as $16$ and training epochs as $200$.
In all experiments, the decoder's learning rate is ten times that of the backbone.
We use the poly scheduling to decay the learning rate during the training process: $lr = lr_{\mathrm{base}} \cdot \left(1 - \frac{\mathrm{iter}}{\mathrm{total\ iter}} \right)^{0.9}$.

\subsection{Comparison with Existing Alternatives}

We compare our method with following recent semi-supervised semantic segmentation methods: Mean Teacher (MT)~\cite{meanteacher}, CCT~\cite{cct},  GCT~\cite{gct}, PseudoSeg~\cite{pseudoseg}, CutMix~\cite{french2019semi}, CPS~\cite{cps}, PC${}^2$Seg~\cite{pc2seg}, AEL~\cite{ael}. 
We re-implement MT~\cite{meanteacher}, CutMix~\cite{cutmix} for a fair comparison.
For Cityscapes~\cite{cityscapes}, we also reproduce CPS~\cite{cps} and AEL~\cite{ael}.
All results are equipped with the same network architecture (DeepLabv3+ as decoder and ResNet-101 as encoder).
It is important to note the \textit{classic} PASCAL VOC 2012 Dataset and \textit{blender} PASCAL VOC 2012 Dataset only differ in training set. 
Their validation set are the same common one with $1,449$ images.

\begin{table}[t]
\centering
\caption{
Comparison with state-of-the-art methods on \textit{blender} \textbf{PASCAL VOC 2012} \texttt{val} set under different partition protocols. 
All labeled images are selected from the augmented VOC \texttt{train} set, which consists of $10,582$ samples in total.
``SupOnly'' stands for supervised training without using any unlabeled data.
\dag\ means we reproduce the approach.
}
\label{tab:blender}
\vspace{-5pt}
\scalebox{0.7}{
\begin{tabular}{l|llll}
\toprule
%  & \multicolumn{4}{c||}{ \textbf{\textit{blender} PASCAL VOC 2012}} &\multicolumn{4}{c}{ \textbf{Cityscapes}} \\
% \midrule
Method & 
1/16 (662) & 1/8 (1323) & 1/4 (2646) & 1/2 (5291) \\
\midrule
SupOnly & 
67.87 & 71.55 & 75.80 & 77.13 \\
\midrule
MT$^\dag$~\cite{meanteacher} & 
70.51 & 71.53 & 73.02 & 76.58  \\
CutMix$^\dag$~\cite{french2019semi} & 
71.66 & 75.51 & 77.33 & 78.21  \\
CCT~\cite{cct} & 
71.86 & 73.68 & 76.51 & 77.40 \\
GCT~\cite{gct} &
70.90 & 73.29 & 76.66 & 77.98 \\
CPS~\cite{cps} & 
74.48 & 76.44 & 77.68 & 78.64\\
AEL~\cite{ael} & 
77.20 & 77.57 & 78.06 & 80.29\\
\midrule
U$^2$PL (w/ CutMix) &
\textbf{77.21} \footnotesize{(\textcolor{blue}{$+$5.55})} &
\textbf{79.01} \footnotesize{(\textcolor{blue}{$+$3.50})} &
\textbf{79.30} \footnotesize{(\textcolor{blue}{$+$1.97})} & 
\textbf{80.50} \footnotesize{(\textcolor{blue}{$+$2.29})} \\
\bottomrule
\end{tabular}
}
\vspace{-5pt}
\end{table}

\begin{table}[t]
\centering
\caption{
Comparison with state-of-the-art methods on \textbf{Cityscapes} \texttt{val} set under different partition protocols. 
All labeled images are selected from the Cityscapes \texttt{train} set, which consists of $2,975$ samples in total.
``SupOnly'' stands for supervised training without using any unlabeled data.
\dag\ means we reproduce the approach.
}
\label{tab:city}
\vspace{-5pt}
\scalebox{0.7}{
\begin{tabular}{l|llll}
\toprule
%  & \multicolumn{4}{c||}{ \textbf{\textit{blender} PASCAL VOC 2012}} &\multicolumn{4}{c}{ \textbf{Cityscapes}} \\
% \midrule
Method & 
1/16 (186) & 1/8 (372) & 1/4 (744) & 1/2 (1488) \\
\midrule
SupOnly & 
65.74 & 72.53 & 74.43 & 77.83 \\
\midrule
MT$^\dag$~\cite{meanteacher} & 
69.03 & 72.06 & 74.20 & 78.15 \\
CutMix$^\dag$~\cite{french2019semi} & 
67.06 & 71.83 & 76.36 & 78.25 \\
CCT~\cite{cct} & 
69.32 & 74.12 & 75.99 & 78.10 \\
GCT~\cite{gct} &
66.75 & 72.66 & 76.11 & 78.34 \\
CPS$^\dag$~\cite{cps} &
69.78 & 74.31 & 74.58 & 76.81 \\
AEL$^\dag$~\cite{ael} &
74.45 & 75.55 & 77.48 & 79.01 \\
\midrule
U$^2$PL (w/ CutMix) &
\textbf{\textit{70.30}} \footnotesize{(\textcolor{blue}{$+$3.24})} & 
\textbf{\textit{74.37}} \footnotesize{(\textcolor{blue}{$+$2.54})} & 
\textbf{\textit{76.47}} \footnotesize{(\textcolor{blue}{$+$0.11})} & 
\textbf{\textit{79.05}} \footnotesize{(\textcolor{blue}{$+$0.80})} \\
U$^2$PL (w/ AEL) &
\textbf{74.90} \footnotesize{(\textcolor{blue}{$+$0.45})} & 
\textbf{76.48} \footnotesize{(\textcolor{blue}{$+$0.93})} & 
\textbf{78.51} \footnotesize{(\textcolor{blue}{$+$1.03})} & 
\textbf{79.12} \footnotesize{(\textcolor{blue}{$+$0.11})} \\
\bottomrule
\end{tabular}
}
\vspace{-15pt}
\end{table}

\noindent\textbf{Results on \textit{classic} PASCAL VOC 2012 Dataset.}
\cref{tab:classic} compares our method with the other state-of-the-art methods on $\textit{classic}$ PASCAL VOC 2012 Dataset.
U$^2$PL outperforms the supervised baseline by $+22.21\%$, $+14.23\%$, $+7.78\%$ and $+4.47\%$ under $1/16$, $1/8$, $1/4$ and $1/2$ partition protocols respectively.
For a fair comparison, we only list the methods tested on \textit{classic} PASCAL VOC 2012. 
Our method U$^2$PL outperform PC${}^2$Seg under all partition protocols by $+10.98\%$, $+2.87\%$, $+3.88\%$ and $+3.11\%$ under $1/16$, $1/8$, $1/4$ and $1/2$ partition protocols respectively.
Even under full supervision, our method outperform PC${}^2$Seg by $+5.34\%$.

\noindent\textbf{Results on \textit{blender} PASCAL VOC 2012 Dataset.} \cref{tab:blender} shows the comparison results on \textit{blender} PASCAL VOC 2012 Dataset. 
Our method U$^2$PL outperforms all the other methods under most partition protocols. 
Compared with the baseline model (trained with only supervised data),  U$^2$PL achieves all improvements of $+9.34\%$, $+7.46\%$, $+3.50\%$ and $+3.37\%$ under $1/16$, $1/8$, $1/4$ and $1/2$ partition protocols respectively.
Compared with the existing state-of-the-art methods, U$^2$PL surpasses them under all partition protocols.
Especially under $1/8$ protocol and $1/4$ protocol, U$^2$PL outperforms AEL by $+1.44\%$ and $+1.24\%$.
%
% Only under the $1/2$ protocol, U$^2$PL is slightly less than AEL by $-0.07\%$, which is almost a negligible gap.

\noindent\textbf{Results on Cityscapes Dataset.} \cref{tab:city} illustrates the comparison results on the Cityscapes \texttt{val} set. 
U$^2$PL achieves consistent performance gains over the supervised only baseline by $+9.16\%$, $+3.95\%$, $+4.08\%$ and $+1.29\%$ under $1/16$, $1/8$, $1/4$ and $1/2$ partition protocols. 
U$^2$PL outperforms the existing state-of-the-art method by a notable margin.
In particular, U$^2$PL outperforms AEL by $+0.45\%$, $+0.93\%$, $+1.03\%$ and $+0.11\%$ under $1/16$, $1/8$, $1/4$ and $1/2$ partition protocols.

Note that when labeled data is extremely limited, \textit{e.g.}, when we only have $92$ labeled data, our U$^2$PL outperforms previous methods by a large margin ($+10.98\%$ under $1/16$ split for classic PASCAL VOC 2012),
proofing the efficiency of using unreliable pseudo-labels.

\subsection{Ablation Studies}

\noindent \textbf{Effectiveness of Using Unreliable Pseudo-Labels.} To prove our core insight, \textit{i.e.}, using unreliable pseudo-labels promotes semi-supervised semantic segmentation,
we conduct experiments about selecting negative candidates (\cref{sec:unreliable}) with different reliability.
\cref{tab:abalation_reliable} demonstrates the mIoU results on PASCAL VOC 2012 \texttt{val} set.
``Unreliable'' outperforms other options, proving using unreliable pseudo-labels does help.
\cref{sec:morecityscapes} shows the effectiveness of using unreliable pseudo-labels on Cityscapes.

\noindent \textbf{Effectiveness of Probability Rank Threshold.} \cref{sec:unreliable} proposes to use probability rank threshold to balance informativeness and confusion caused by unreliable pixels.
\cref{tab:abalation_probability_rank} provides a verification that such balance promotes the performance.
$r_l = 3$ and $r_h = 20$ outperform other options by a large margin.
When  $r_l = 1$, false negative candidates would not be filtered out, causing the intra-class features of pixels incorrectly distinguished by $\mathcal{L}_{c}$. 
When  $r_l = 10$, negative candidates tend to become irrelevant with corresponding anchor pixels in semantic, making such discrimination less informative.
\cref{sec:hyper_city} studies PRT and $\alpha_0$ on Cityscapes.

\begin{table}[t]
\centering
\caption{
\textbf{Ablation study on using pseudo pixels with different reliability}, which is measured by the entropy of pixel-wise prediction (see \cref{sec:unreliable}). 
``Unreliable'' denotes selecting negative candidates from pixels with top 20\% highest entropy scores.
``Reliable'' denotes the bottom 20\% counterpart. 
``All'' denotes sampling regardless of entropy.
%
% We prove this effectiveness under 1/4 and 1/8 partition protocol on \textit{blender} PASCAL VOC 2012 Dataset.
}
\label{tab:abalation_reliable}
\vspace{-5pt}
\setlength{\tabcolsep}{10pt}
\begin{tabular}{cccc}
\toprule
 & Unreliable & Reliable & All  \\
\midrule
1/8 (1323) & \textbf{79.01} & 77.30 & 77.40 \\
1/4 (2646) & \textbf{79.30} & 77.35 & 77.57 \\
% \checkmark         &           &           &         &             \\
\bottomrule
\end{tabular}
\vspace{-5pt}
\end{table}

\begin{table}[t]
\centering
\caption{
\textbf{Ablation study on the probability rank threshold}, which is described in \cref{sec:unreliable}.
}
\label{tab:abalation_probability_rank}
\vspace{-5pt}
\setlength{\tabcolsep}{16pt}
\begin{tabular}{cccc}
\toprule
  $r_l$ & $r_h$ &  1/8 (1323) & 1/4 (2646)  \\
\midrule
 1 & 3&78.57  &79.03\\
 1 & 20&78.64  &79.07\\
 3 & 10 &78.27  &78.91\\
 3 & 20 &\textbf{79.01}  &\textbf{79.30} \\
 10 & 20 &78.62 &78.94\\
% \checkmark         &           &           &         &             \\
\bottomrule
\end{tabular}
\vspace{-5pt}
\end{table}

\begin{table}[t]
\centering
\caption{
\textbf{Ablation study on the effectiveness of various components in our U$^2$PL}, including unsupervised loss $\mathcal{L}_u$, contrastive loss $\mathcal{L}_c$, category-wise memory bank $\mathcal{Q}_c$, Dynamic Partition Adjustment (DPA), Probability Rank Threshold (PRT), and high entropy filtering (Unreliable).
}
\label{tab:abalation_dynamic}
\vspace{-5pt}
\setlength{\tabcolsep}{7.5pt}
\begin{tabular}{ccccc|c}
\toprule
 $\mathcal{L}_{c}$ & $\mathcal{Q}_c$ & DPA & PRT & Unreliable & 1/4 (2646) \\
\midrule
 & & & & &  73.02\\
\checkmark & & & & &77.08  \\
\checkmark & \checkmark & &\checkmark &\checkmark & 78.49  \\
\checkmark & \checkmark & \checkmark & &\checkmark & 79.07 \\
 \checkmark & \checkmark & \checkmark & \checkmark & & 77.57 \\
 \checkmark & \checkmark & \checkmark & \checkmark & \checkmark & \textbf{79.30}\\
\bottomrule
\end{tabular}
\vspace{-5pt}
\end{table}

\begin{table}[t]
\centering
\caption{
\textbf{Ablation study on $\alpha_0$ in \cref{eq:dpa}}, which controls the initial proportion between reliable and unreliable pixels.
}
\setlength{\tabcolsep}{10pt}
\label{tab:abalation_alpha}
\vspace{-5pt}
\begin{tabular}{ccccc}
\toprule
$\alpha_0$    &      40\% & 30\% & 20\% & 10\%   \\
\midrule
1/8 (1323)   &  76.77     &77.34           &\textbf{79.01}            & 77.80           \\
1/4 (2646)   &  76.92     &77.38           &\textbf{79.30}            & 77.95           \\
\bottomrule
\end{tabular}
\vspace{-15pt}
\end{table}

\begin{figure*}[t]
    \centering
    \includegraphics[width=1.0\textwidth]{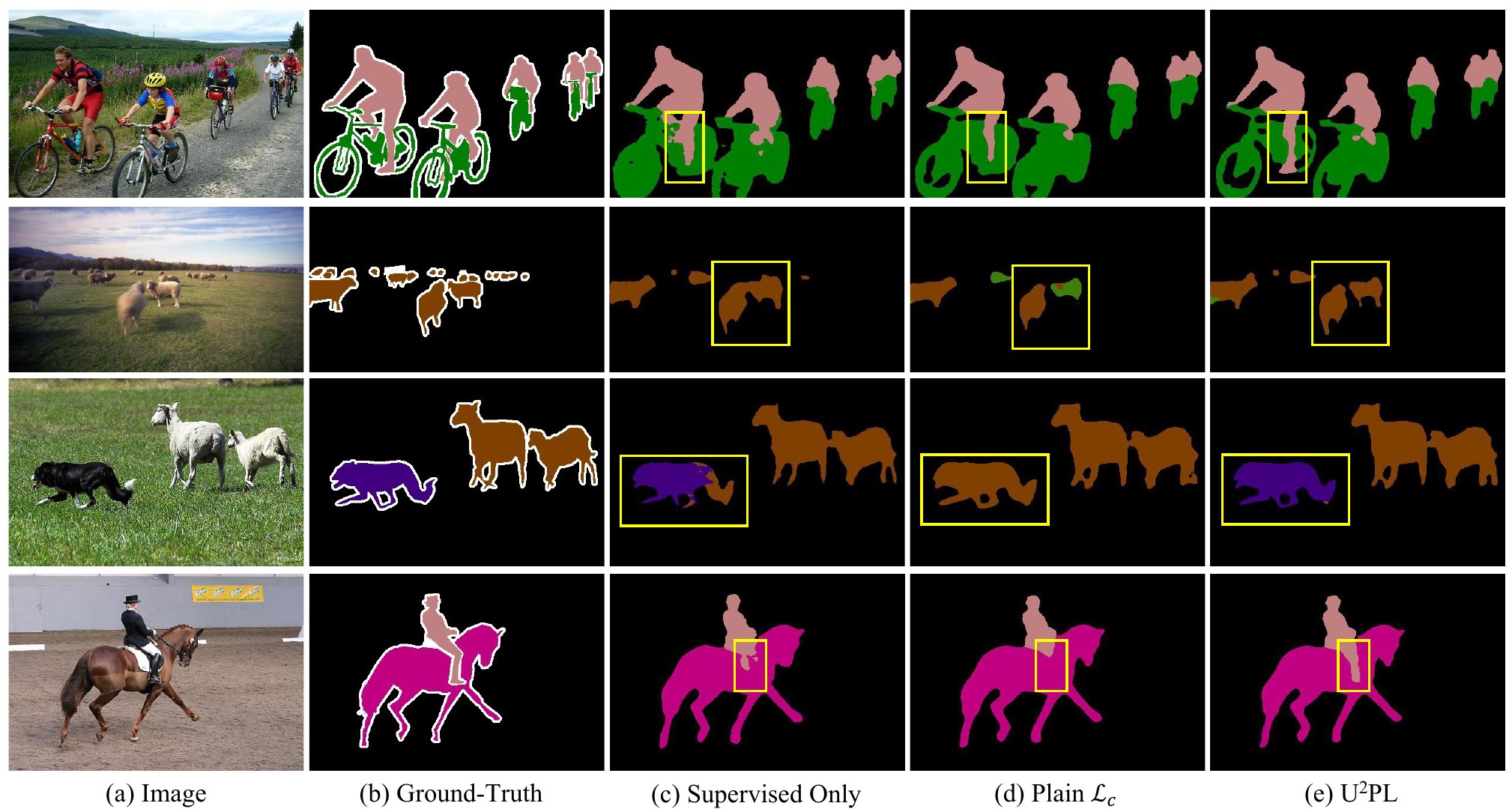}
    \vspace{-18pt}
    \caption{
    Qualitative results on \textbf{PASCAL VOC 2012} \texttt{val} set.
    All models are trained under the $1/4$ partition protocol of \textit{blender} set, which contains $2,646$ labeled images and $7,396$ unlabeled images.
    (a) Input images. 
    (b) Hand-annotated labels for the corresponding image. 
    (c) \textit{Only} labeled images are used for training without any unlabeled data.
    (d) The vanilla contrastive learning framework, where all pixels are used as negative samples without entropy filtering.
    (e) Predictions from our U${}^2$PL.
    Yellow rectangles highlight the promotion of segmentation results by adequately using unreliable pseudo-labels.
    }
    \label{fig:visual}
    \vspace{-15pt}
\end{figure*}

\noindent \textbf{Effectiveness of Components.} We conduct experiments in \cref{tab:abalation_dynamic} to ablate each component of U$^2$PL step by step. 
For a fair comparison, all the ablations are under 1/4 partition protocol on blender PASCAL VOC 2012 Dataset. 
Above all, we use no $\mathcal{L}_{c}$ trained model as our baseline, achieving mIoU of $73.02\%$ (MT in \cref{tab:blender}).
Simply adding $\mathcal{L}_{c}$ without DPA strategy improves the baseline by $+4.06\%$.
Category-wise memory bank $\mathcal{Q}_{c}$, along with PRT and high entropy filtering brings an improvement by $+5.47\%$ to baseline.
Dynamic Partition Adjustment (DPA) together with high entropy filtering, brings an improvement by $+6.05\%$ to baseline.
Note that DPA is a linear adjustment without tuning (refer to \cref{eq:dpa}), which is simple yet efficient.
For Probability Rank Threshold (PRT) component, we set corresponding parameter according to \cref{tab:abalation_probability_rank}. 
Without high entropy filtering, the improvement decreased significantly at $+4.55\%$.
Finally, when adding all the contribution together, our method achieves state-of-the-art result under $1/4$ partition protocol with mIoU of $79.30\%$.
Following this result, we apply these components and corresponding parameters in all experiments on \cref{tab:blender} and \cref{tab:classic}.

\noindent \textbf{Ablation Study on Hyper-parameters.}
We ablate following important parameter for U$^2$PL. 
\cref{tab:abalation_alpha} studies the impact of different initial reliable-unreliable partition.
This parameter $\alpha_0$ have a certain impact on performance. We find $\alpha_0 = 20\%$ achieves the best performance.
Small $\alpha_0$ will introduce incorrect pseudo labels for supervision, and large $\alpha_0$ will make the information of some high-confidence samples underutilized.
Other hyper-parameters are studied in \cref{sec:hyper_voc}.

% \begin{table}[htbp]
% \centering
% \caption{
% %
% Study on weight $\lambda_c$ for balancing $\mathcal{L}_{c}$.
% %
% }
% \label{tab:abalation_lambda}

% \begin{tabular}{cccccc}
% \toprule
% $\lambda_c$    &      10 &5& 1 & 0.5 & 0.1   \\
% \midrule
% 1/8 (1323)   &          &       &           &            &            \\
% 1/4 (2646)   &          &       &           &            &             \\
% \bottomrule
% \end{tabular}
% \end{table}

\subsection{Qualitative Results}

\cref{fig:visual} shows the results of different methods on the PASCAL VOC 2012 \texttt{val} set.
Benefiting from using unreliable pseudo-labels, U${}^2$PL outperforms other methods.
Note that using contrastive learning without filtering those unreliable pixels, sometimes does harm to the model (see row 2 and row 4 in \cref{fig:visual}), leading to worse results than those when the model is trained only by labeled data.

Furthermore, through visualizing the segmentation results, we find that our method achieves much better performance on those ambiguous regions (\textit{e.g.}, the border between different objects).
Such visual difference proves that our method finally makes the reliability of unreliable prediction labels stronger.

%%%% Section: Conclusion
\section{Conclusion}\label{sec:conclusion}
We propose a semi-supervised semantic segmentation framework U$^2$PL by including unreliable pseudo-labels into training,
%
% Our main insight is that those unreliable ones are just confused about a few classes and are confident enough for not belonging to the remaining remote classes.
%
which outperforms many existing state-of-the-art methods, suggesting our framework provide a new promising paradigm in semi-supervised learning research.
Our ablation experiments proves the insight of this work is quite solid.
Qualitative result gives a visual proof for its effectiveness, especially the better performance on borders between semantic objects or other ambiguous regions.

The training of our method is time-consuming compared with fully-supervised methods~\cite{fcn, unet, deeplabv3p, deeplab, pspnet}, which is a common disadvantage for semi-supervised learning tasks~\cite{ael, dars, st++, cps, cct, pc2seg}.
Due to the extreme lack of labels, the semi-supervised learning frameworks commonly need to pay a price in time for higher accuracy.
More in-depth exploration could be conducted on their training optimization in the future.

%%%% References
{\small
\bibliographystyle{ieee_fullname}
\bibliography{ref}
}

%%%% Appendix
\appendix
\renewcommand\thefigure{A\arabic{figure}}
\renewcommand\thetable{A\arabic{table}}  
\renewcommand\theequation{A\arabic{equation}}
\setcounter{equation}{0}
\setcounter{table}{0}
\setcounter{figure}{0}

\section*{Appendix}

We organize the Appendix as follows.
Above all, more details for reproducing the results will be given in \cref{sec:reproduce}.
Then we will give more results on Cityscapes from two perspectives in \cref{sec:morecityscapes}.
We also provide an alternative of contrastive learning to prove our main insight does not only rely on contrastive learning in \cref{sec:alternative}.
Besides, ablation studies on both PASCAL VOC 2012 and Cityscapes for more hyper-parameters are given in \cref{sec:hyper}.
Finally, visualization on feature space gives a visual proof for the effectiveness of U${}^2$PL in \cref{sec:featurespace}.

\begin{figure*}[t]
    \centering
    \includegraphics[width=1\textwidth]{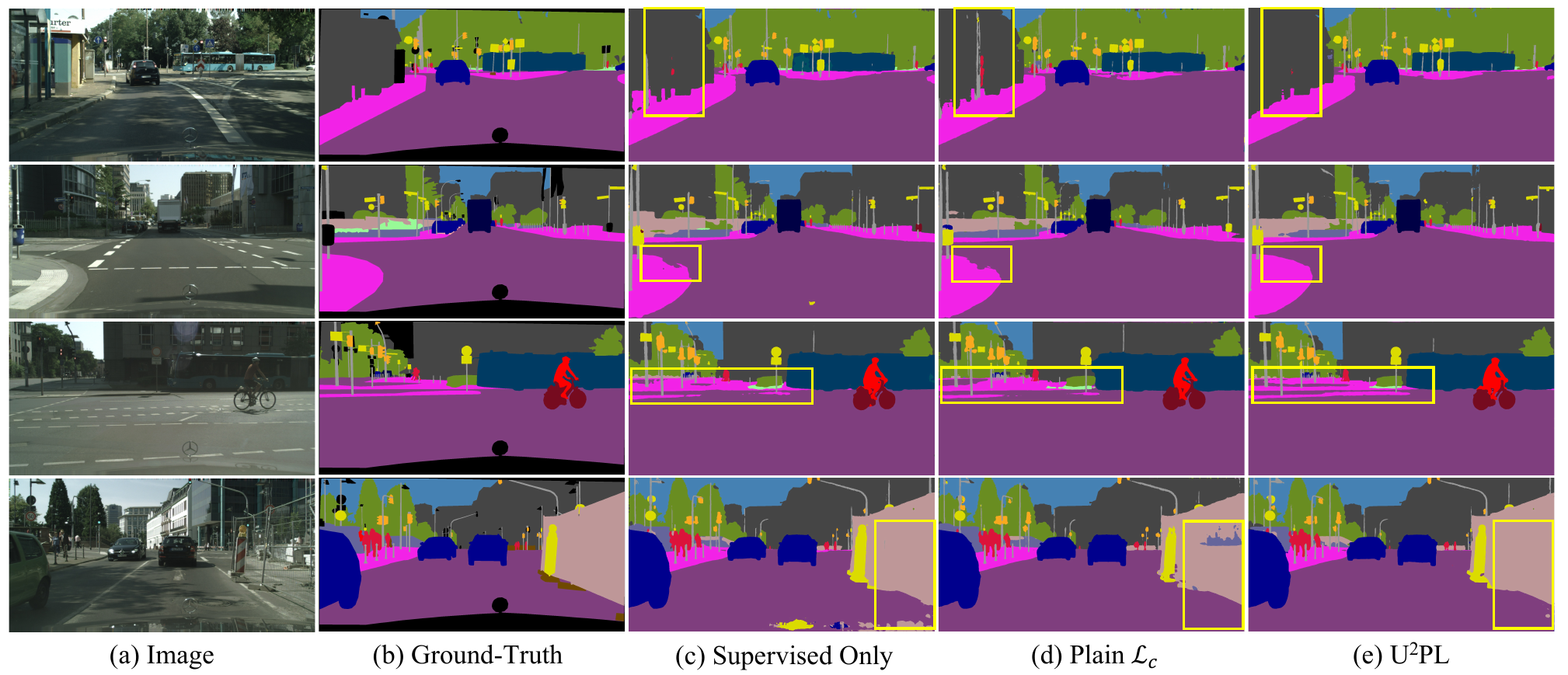}
    \vspace{-18pt}
    \caption{
    Qualitative results on \textbf{Cityscapes} \texttt{val} set.
    All models are trained under the $1/2$ partition protocol, which contains $1,488$ labeled images and $1,487$ unlabeled images.
    (a) Input images. 
    (b) Hand-annotated labels for the corresponding image. 
    (c) \textit{Only} labeled images are used for training.
    (d) The vanilla contrastive learning framework, where all pixels are used as negative samples without entropy filtering.
    (e) Predictions from our U${}^2$PL.
    Yellow rectangles highlight the promotion by adequately using unreliable pseudo-labels.
    }
    \label{fig:visual_city}
    \vspace{-10pt}
\end{figure*}

\section{More Details for Reproducibility}\label{sec:reproduce}

For Cityscapes~\cite{cityscapes}, we utilize OHEM which is the same as previous methods~\cite{cps, ael}.
The temperature $\tau$ is set to $0.5$ for both PASCAL VOC 2012~\cite{voc} and Cityscapes~\cite{cityscapes}.
We use SGD optimizer for all experiments.
For experiments in PASCAL VOC 2012~\cite{voc}, the initial base learning rate is $0.001$ and the weight decay is $0.0001$.
For experiments in Cityscapes~\cite{cityscapes}, the initial base learning rate is $0.01$ and the weight decay is $0.0005$.
In our experiments, we find if we train the model only with supervised loss for the initial a few epochs then apply U$^2$PL, it can achieve better performance.
We define such epoch as the warm start epoch, and the corresponding warm start epochs for PASCAL VOC 2012 and Cityscapes are $1$ and $20$ respectively.

To prevent overfitting, we apply random cropping, random horizontal flipping, and random scaling with the range of $[0.5, 2.0]$ for both PASCAL VOC 2012~\cite{voc} and Cityscapes~\cite{cityscapes} following previous methods~\cite{ael,cps,pc2seg,pseudoseg}.
Our memory queue is category-specific. 
For the background category, the length of the queue is set to be $50,000$.
For foreground categories, the length of the queue is all $30,000$.
All baselines \textit{i.e.}, ``SupOnly'', ``MT'', and ``CutMix'' are re-implemented by ourselves,
where the only difference between ``MT'' and ``CutMix'' is that the latter applies CutMix~\cite{cutmix} augmentation for unlabeled images.

The hyper-parameters used in this work are listed in \cref{tab:hyper}. Among them, $M, N, \delta_p$ are used for contrastive learning, for which we simply follow \cite{reco}. $\lambda_c, \eta, \tau$ are training-related, while $\alpha_0, r_l, r_h$ are additionally introduced by our U$^2$PL.

\vspace{-5pt}
\begin{table}[!ht]
    \setlength{\tabcolsep}{7pt}
    \centering
    \caption{
        \textbf{Summary of hyper-parameters} used in U$^2$PL.
    }
    \label{tab:hyper}
    \vspace{-8pt}
    \scalebox{0.8}{
    \begin{tabular}{l l l}
        \toprule
        Symbol & Description & Default Value \\
        \midrule
        $(M, N)$ & contrastive learning settings & (50, 256) \\
        $\delta_p$ & confidence threshold of positive samples & 0.3 \\
        $(\lambda_c, \eta)$ & loss weights & (0.1, 1) \\
        $\tau$ & loss temperature & 0.5 \\
        $\alpha_0$ & initial proportion of unreliable pixels & 20\% \\
        $(r_l, r_h)$ & probability rank thresholds & (3, 20) \\
        \bottomrule
    \end{tabular}
    }
    \vspace{-5pt}
\end{table}

\begin{figure*}[t]
    \centering
    \includegraphics[width=1.0\textwidth]{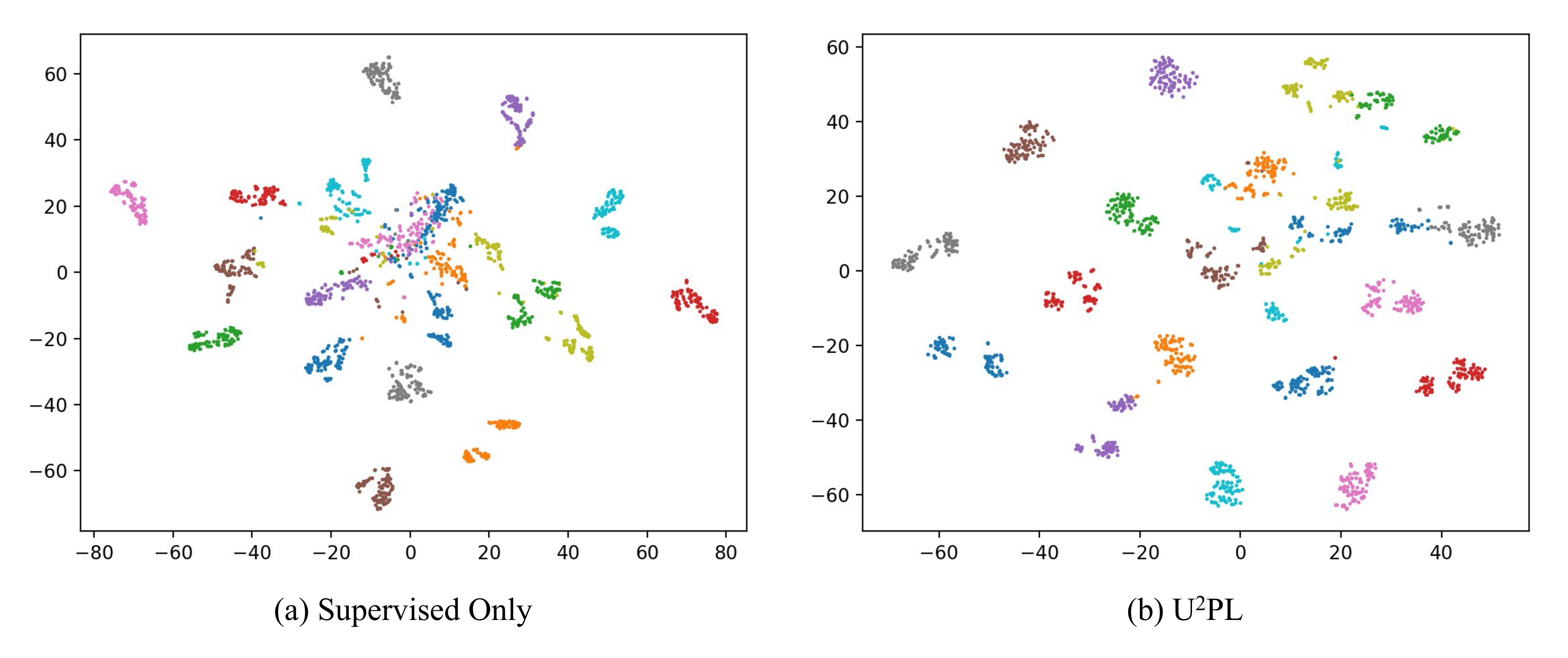}
    \vspace{-25pt}
    \caption{
        \textbf{Visualization of the feature spaces} learned by our U${}^2$PL and its supervised counterpart, using t-SNE~\cite{tsne}.
        The training set is the 1/4 partition protocol (2646) in \textit{blender} VOC PASCAL 2012 Dataset.
    }
    \label{fig:tsne}
    \vspace{-5pt}
\end{figure*}

\section{More Results on Cityscapes}
\label{sec:morecityscapes}

\noindent \textbf{Quantitative Results.} \cref{tab:abalation_reliable_city} demonstrates the mIoU results on Cityscapes \texttt{val} set.
``Unreliable'' outperforms other options, proving using unreliable pseudo-labels does help.
U$^2$PL fully mines the information of all pixels.

\begin{table}[t]
\centering
\caption{
\textbf{Ablation study on using pseudo pixels with different reliability}, which is measured by the entropy of pixel-wise prediction. 
``Unreliable'' denotes selecting negative candidates from pixels with top 20\% highest entropy scores.
``Reliable'' denotes the bottom 20\% counterpart. 
``All'' denotes sampling regardless of entropy.
We prove this effectiveness under $1/2$ and $1/4$ partition protocol on Cityscapes \texttt{val} set.
}
\label{tab:abalation_reliable_city}
\vspace{-5pt}
\setlength{\tabcolsep}{12pt}
\begin{tabular}{c|ccc}
\toprule
 & Unreliable & Reliable & All  \\
\midrule
1/2 (1488) & \textbf{79.05} & 77.19 & 76.96 \\
1/4 (744) & \textbf{76.47} & 75.16 & 74.51 \\
% \checkmark         &           &           &         &             \\
\bottomrule
\end{tabular}
\vspace{-10pt}
\end{table}

\noindent \textbf{Qualitative Results.} \cref{fig:visual_city} shows the results of different methods on the Cityscapes \texttt{val} set.
Benefiting by using unreliable pseudo-labels, U${}^2$PL outperforms other methods.
Note that using contrastive learning without filtering those unreliable pixels, sometimes does harm to the model (see the $1$-st row and the $4$-th row in \cref{fig:visual_city}), leading to worse results than those when the model is trained only by labeled data.
Such visual difference proves that our method finally makes the reliability of unreliable prediction labels stronger.

\section{Alternative of Contrastive Learning}
\label{sec:alternative}

Our proposed U${}^2$PL is not limited by contrastive learning.
Binary classification is also a sufficient way to use unreliable pseudo-labels, \textit{i.e.}, using binary cross-entropy loss (BCE) $\mathcal{L}_b$ other than contrastive loss.
For $i$-th anchor $\mathbf{z}_{ci}$ belongs to class $c$, we simply use its negative samples $\{\mathbf{z}_{cij}^{-}\}_{j=1}^{N}$ and positive sample $\mathbf{z}_c^{+}$ to compute the BCE loss:
\begin{equation}
\label{eq:binary}
\begin{aligned}
    \mathcal{L}_b = -&\frac{1}{C\times M\times N}\sum_{c=0}^{C-1} \sum_{i=1}^M \sum_{j=1}^N \\ 
    &\log\left[ \frac{e^{\langle \mathbf{z}_{ci}, \mathbf{z}_{c}^{+} \rangle / \tau}}{e^{\langle \mathbf{z}_{ci}, \mathbf{z}_{c}^{+} \rangle / \tau} + e^{\langle \mathbf{z}_{ci}, \mathbf{z}_{cij}^{-} \rangle / \tau}} \right],
\end{aligned}
\end{equation}
where $C$, $M$, and $N$ are the total number of classes, anchor pixels, and negative samples, respectively.
$\langle\cdot, \cdot\rangle$ is the cosine similarity of two features, and $\tau$ represents the temperature.

\cref{tab:city_bce} and \cref{tab:blender_bce} are results of using unreliable pseudo-labels based on binary classification on Cityscapes~\cite{cityscapes} and PASCAL VOC 2012~\cite{voc} \texttt{val} set respectively.
From \cref{tab:city_bce} and \cref{tab:blender_bce}, we can tell that our U${}^2$PL is not restricted by contrastive learning, a basic binary classification also does help.
On Cityscapes \texttt{val} set, U${}^2$PL with $\mathcal{L}_b$ can outperforms supervised only baseline by $+3.77\%$, $+0.40\%$, $+1.48\%$, and $+0.53\%$ under $1/16$, $1/8$, $1/4$, and $1/2$ partial protocols.
U${}^2$PL with $\mathcal{L}_b$ can outperforms supervised only baseline by $+7.49\%$, $+5.07\%$, $+3.84\%$, and $+2.67\%$ under $1/16$, $1/8$, $1/4$, and $1/2$ partial protocols on PASCAL VOC 2012 \texttt{val} set.
Note that under the $1/4$ partition protocol of \textit{blender} PASCAL VOC 2012, the bianry classification based U${}^2$PL (w/ $\mathcal{L}_b$) outperforms the contrastive learning based U${}^2$PL (w/ $\mathcal{L}_c$) by $+0.34\%$, which proves that contrastive learning is not the only efficient way of using unreliable pseudo-labels.

\begin{table}[t]
\centering
\caption{
Using unreliable pseudo-labels based on binary classification on \textbf{Cityscapes} \texttt{val} set under different partition protocols. 
}
\label{tab:city_bce}
\vspace{-8pt}
\scalebox{0.82}{
\begin{tabular}{l|cccc }
\toprule
%  & \multicolumn{4}{c||}{ \textbf{\textit{blender} PASCAL VOC 2012}} &\multicolumn{4}{c}{ \textbf{Cityscapes}} \\
% \midrule
Method & 
1/16 (186) & 1/8 (372) & 1/4 (744) & 1/2 (1488) \\
\midrule
SupOnly & 
65.74 & 72.53 & 74.43 & 77.83 \\
MT~\cite{meanteacher} & 
69.03 & 72.06 & 74.20 & 78.15 \\
\midrule
U$^2$PL (w/ $\mathcal{L}_c$) &
\textbf{70.30} & \textbf{74.37}  &\textbf{76.47}  &  \textbf{79.05} \\
U$^2$PL (w/ $\mathcal{L}_b$) &
69.87 & 72.93 & 75.91 & 78.36 \\
\bottomrule
\end{tabular}
}
\vspace{-5pt}
\end{table}

\begin{table}[t]
\centering
\caption{
Using unreliable pseudo-labels based on binary classification on \textbf{PASCAL VOC 2012} \texttt{val} set under different splits. 
}
\label{tab:blender_bce}
\vspace{-8pt}
\scalebox{0.82}{
\begin{tabular}{l|cccc }
\toprule
%  & \multicolumn{4}{c||}{ \textbf{\textit{blender} PASCAL VOC 2012}} &\multicolumn{4}{c}{ \textbf{Cityscapes}} \\
% \midrule
Method & 
1/16 (662) & 1/8 (1323) & 1/4 (2646) & 1/2 (5291) \\
\midrule
SupOnly & 
67.87 & 71.55 & 75.80 & 77.13 \\
MT~\cite{meanteacher} & 
70.51 & 71.53 & 73.02 & 76.58  \\
\midrule
U$^2$PL (w/ $\mathcal{L}_c$) &
\textbf{77.21} & \textbf{79.01}  & 79.30 & \textbf{80.50} \\
U$^2$PL (w/ $\mathcal{L}_b$) &
75.36 & 76.62  & \textbf{79.64} & 79.80 \\
\bottomrule
\end{tabular}
}
\vspace{-5pt}
\end{table}

\section{More Ablation Studies}
\label{sec:hyper}

\subsection{More Hyper-parameters on VOC}
\label{sec:hyper_voc}
\noindent \textbf{Base Learning Rate.}
The impact of the base learning rate is shown in \cref{tab:baselr}. 
Results are based on  U${}^2$PL on \textit{blender} VOC PASCAL 2012 Dataset. 
We find that 0.001 outperforms other alternatives.

\noindent \textbf{Temperature.}
\cref{tab:temperature} gives a study on the effect of temperature $\tau$.
Temperature $\tau$ plays an important role to adjust the importance to hard samples
When $\tau=0.5$, our U${}^2$PL achieves best results.
Too large or too small of $\tau$ will have an adverse effect on overall performance.

\subsection{Ablation Studies on Cityscapes}
\label{sec:hyper_city}
\noindent\textbf{Probability Rank Threshold.} 
\cref{tab:prt_city} provides a verification that such balance promotes the performance. $r_l=3$ and $r_h=20$ outperform other options by a large margin.

\noindent\textbf{Initial Reliable-Unreliable Partition.} 
\cref{tab:alpha_city} studies the impact of different $\alpha_0$.
When $\alpha_0=20\%$, the model achieves the best performance.

\begin{table}[t]
\centering
\caption{
\textbf{Ablation study on base learning rate} under 1/4 partition protocol (2646) in \textit{blender} VOC PASCAL 2012 Dataset.
}
\setlength{\tabcolsep}{8.5pt}
\label{tab:baselr}
\vspace{-8pt}
\begin{tabular}{c|ccccc }
\toprule
$lr_{\mathrm{base}}$& 
$10^{-1}$ & $10^{-2}$ & $10^{-3}$ & $10^{-4}$ &  $10^{-5}$ \\
\midrule
mIoU 
& 3.49 & 77.82 &\textbf{79.30} &74.58  &65.69 \\
\bottomrule
\end{tabular}
\vspace{-5pt}
\end{table}

\begin{table}[t]
\centering
\caption{
\textbf{Ablation study on temperature} under 1/4 partition protocol (2646) in \textit{blender} VOC PASCAL 2012 Dataset.
}
\setlength{\tabcolsep}{8.4pt}
\label{tab:temperature}
\vspace{-8pt}
\begin{tabular}{c|ccccc }
\toprule
$\tau$& 
$10$ & $1$ & $0.5$ & $0.1$ &  $0.01$ \\
\midrule
mIoU & 
78.88 & 78.91 &\textbf{79.30} &79.22  &78.78 \\
\bottomrule
\end{tabular}
\vspace{-5pt}
\end{table}

\begin{table}[t]
    \setlength{\tabcolsep}{7pt}
    \centering
    \caption{
        \textbf{Ablation study on PRT} on Cityscapes \texttt{val} set.
    }
    \label{tab:prt_city}
    \vspace{-8pt}
    \scalebox{1}{
    \begin{tabular}{cccccc}
        \toprule
        $r_l$ & 1 & 1 & 3 & 3 & 10 \\
        $r_h$ & 3 & 20 & 10 & 20 & 20 \\
        \midrule
        1/8 (372) & 71.41 & 72.08 & 72.60 & \textbf{74.37} & 72.24 \\
        1/4 (744) & 76.27 & 76.04 &76.01 & \textbf{76.47} & 76.18 \\
        \bottomrule
    \end{tabular}
    }
    \vspace{-5pt}
\end{table}

\begin{table}[t]
    \setlength{\tabcolsep}{10.5pt}
    \centering
    \caption{
        \textbf{Ablation study on $\alpha_0$} on Cityscapes \texttt{val} set.
    }
    \label{tab:alpha_city}
    \vspace{-8pt}
    \scalebox{1}{
    \begin{tabular}{ccccc}
        \toprule
        $\alpha_0$ & 40\% & 30\% & 20\% & 10\%\\
        \midrule
        1/8 (372) & 72.07 & 72.93 &\textbf{74.37} & 71.63 \\
        1/4 (744) & 75.20 & 76.08 &\textbf{76.47} & 76.40 \\
        \bottomrule
    \end{tabular}
    }
    \vspace{-5pt}
\end{table}

\section{Visualization on Feature Space}
\label{sec:featurespace}

To have a better understanding of U${}^2$PL, we give an illustration on visualization of feature space.
Two t-SNE~\cite{tsne} plots are given respectively on the supervised only method and U${}^2$PL.

We can observe from \cref{fig:tsne} that decision boundaries of features generated by the supervised only method are quite confusing, while U${}^2$PL has much more clear ones.
This explains why U${}^2$PL works from a feature point of view.

\end{document}